\documentclass[twoside,twocolumn]{article}
\usepackage{graphicx}
\def\subsection#1{\vspace{0.5ex plus 1ex minus 0.5ex}\noindent{\bf #1.}}

\begin{document}

\title{\vspace*{-2cm}
   {\normalsize Technical Report IDSIA-04-04 \hfill 20 March 2004\\[1mm]}
  \huge\sc\hrule height1pt \vskip 2mm
     \centerline{Tournament versus Fitness Uniform Selection}
     \vskip 2mm \hrule height1pt}

\author{\normalsize\hspace{-1cm}
\begin{minipage}{6cm}
\begin{center}
  Shane Legg \\
  IDSIA\\
  Galleria 2, CH-6928,\\
  Manno-Lugano, Switzerland\\
  Email: shane@idsia.ch
\end{center}
\end{minipage}
\hfil
\begin{minipage}{6cm}
\begin{center}
  Marcus Hutter\\
  IDSIA\\
  Galleria 2, CH-6928,\\
  Manno-Lugano, Switzerland\\
  Email: marcus@idsia.ch
\end{center}
\end{minipage}
\hfil
\begin{minipage}{6cm}
\begin{center}
  Akshat Kumar\\
  Indian Institute of Technology\\
  Guwahati\\
  India\\
  Email: akshat@iitg.ernet.in
\end{center}
\end{minipage}
}

\date{}

\maketitle              % typeset the title of the contribution

\begin{abstract}
\footnote{This work was supported by SNF grant 2100-67712.02.}%
In evolutionary algorithms a critical parameter that must be tuned is
that of selection pressure.  If it is set too low then the rate of
convergence towards the optimum is likely to be slow.  Alternatively
if the selection pressure is set too high the system is likely to
become stuck in a local optimum due to a loss of diversity in the
population. The recent Fitness Uniform Selection Scheme (FUSS) is a
conceptually simple but somewhat radical approach to addressing this
problem --- rather than biasing the selection towards higher fitness,
FUSS biases selection towards sparsely populated fitness levels. In
this paper we compare the relative performance of FUSS with the well
known tournament selection scheme on a range of problems.
\end{abstract}

%%%%%%%%%%%%%%%%%%%%%%%%%%%%%%%%%%%%%%%%%%%%%%%%%%%%%%%%%%%%%%%
\section{Introduction}\label{secIntro}
%%%%%%%%%%%%%%%%%%%%%%%%%%%%%%%%%%%%%%%%%%%%%%%%%%%%%%%%%%%%%%%

The standard selection schemes used in evolutionary algorithms (such
as tournament, ranking, proportional, truncation selection, and so on)
all focus the selection pressure towards individuals of higher fitness
in the population.  The rational being that these individuals are the
most likely to produce offspring (either by mutation or crossover or
both) that belong to still higher fitness levels.  For many problems
this is often a valid assumption, however for difficult, deceptive and
highly multi-modal functions the path towards the global optimum is
rarely smooth (by ``deceptive'' we mean in the general rather than the
technical sense).  In these cases it is important that we explore the
solution space very carefully before becoming too committed to any
subset of solutions that appears to be promising.  In order to do this
we must not focus too much of our search energy on only the most fit
individuals.  In particular we must ensure that we keep some less fit
individuals in the population in case we need to use them to
initialize a new direction of exploration should we become stuck in a
local optimum.

For standard selection schemes this is controlled by appropriately
setting the parameters that govern the selection pressure on the
individuals.  If this pressure is set too high the evolutionary
algorithm (EA) will converge quickly but possibly to a local optimum,
while if it is set too low the system will converge only very slowly,
if at all.  It is often the case that this can only be done through a
process of experimentation with the particular problem at hand.

Many systems have been devised to help prevent this problem by
ensuring that the population maintains a certain degree of diversity.
Significant contributions in this direction are fitness sharing
\cite{Goldberg:87}, crowding \cite{DeJong:75} and local mating
\cite{Collins:91}.  The Fitness Uniform Selection Scheme (FUSS)
\cite{Hutter:01fuss} is another proposed solution to this problem
which is well suited for very difficult optimization problems.  The
key idea is to preserve genetic diversity in the population by using
the fitness of individuals to estimate their similarity.  The beauty
of this approach is that it is simple to implement, problem intrinsic
and also representation independent.

In this paper we present the first experimental investigations into
the performance of FUSS.  Our goal is to develop a better
understanding of its performance characteristics in practice and in
particular how it compares to a standard selection scheme which favors
fit individuals.

\begin{figure*}
\includegraphics[width=0.32\textwidth]{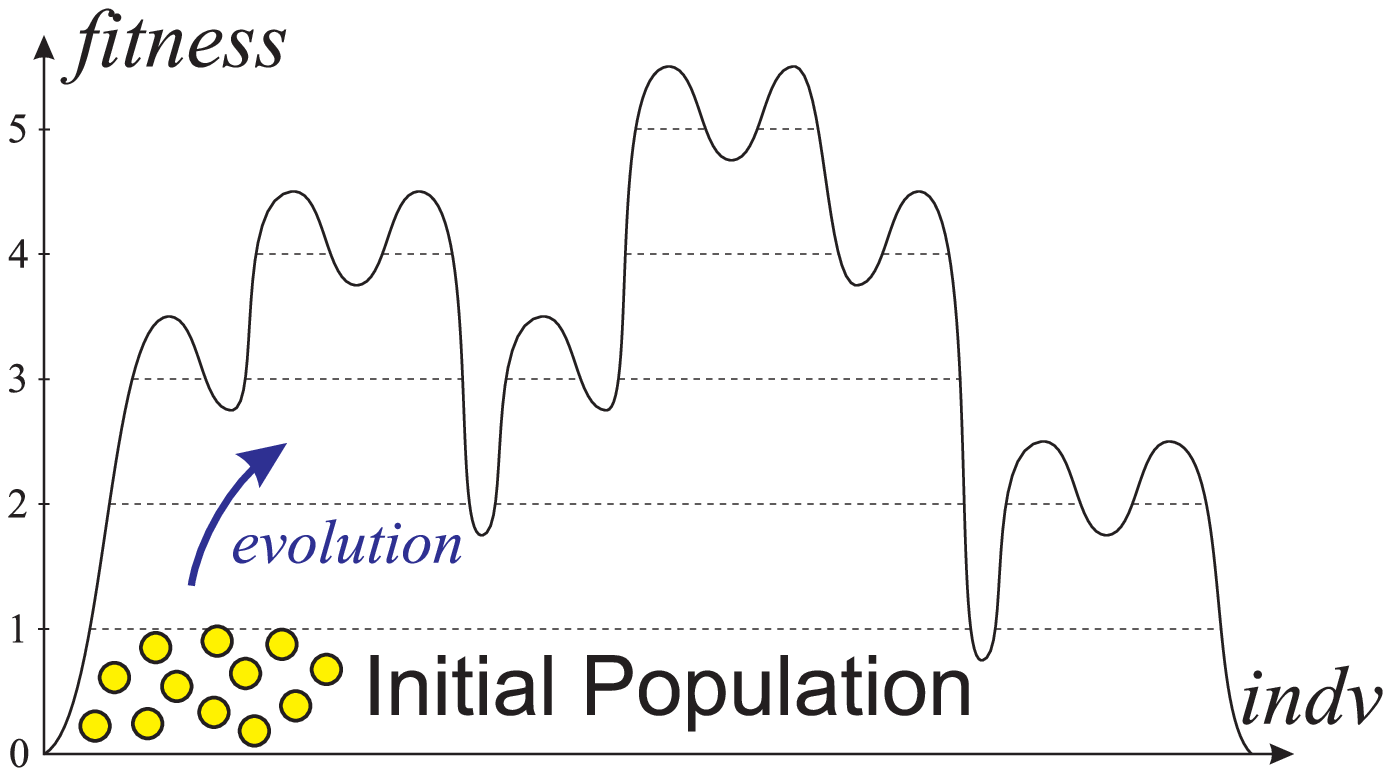}
\includegraphics[width=0.32\textwidth]{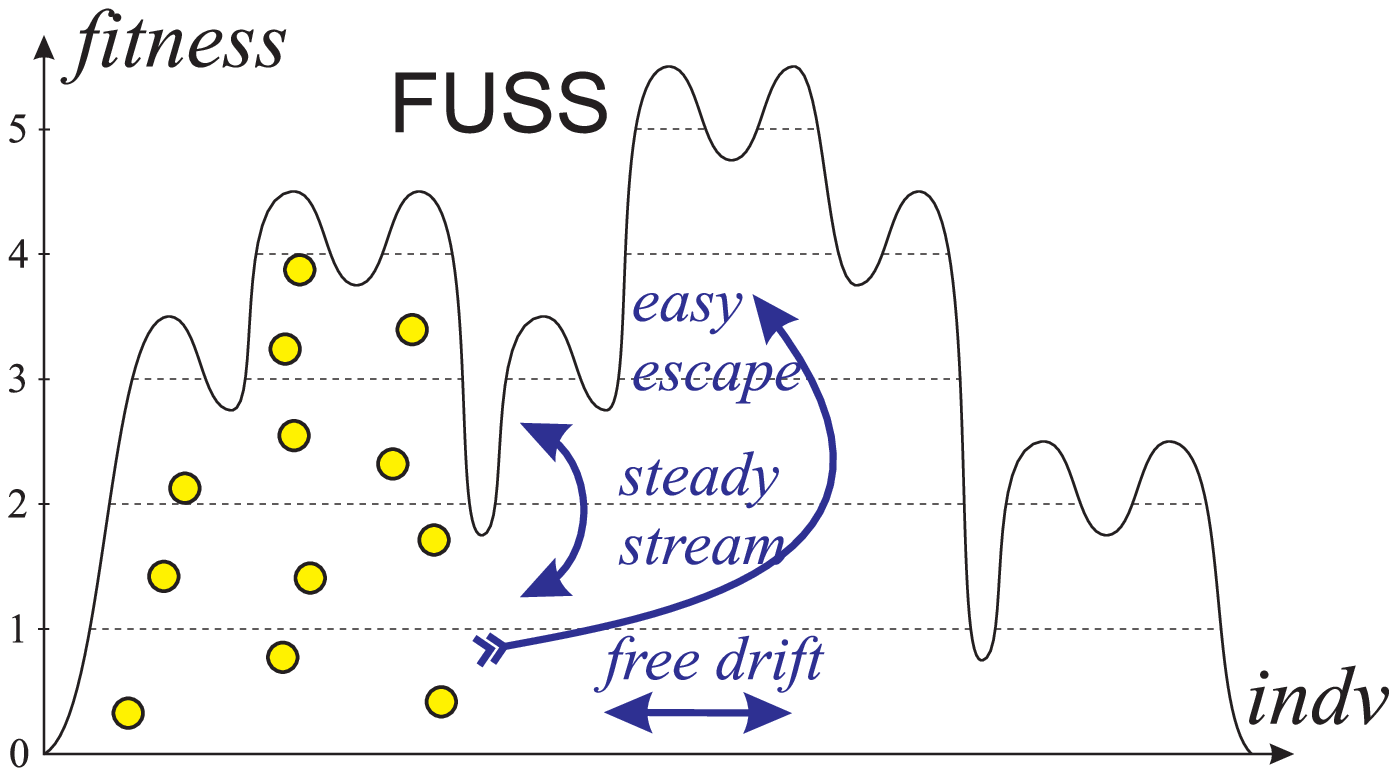}
\includegraphics[width=0.32\textwidth]{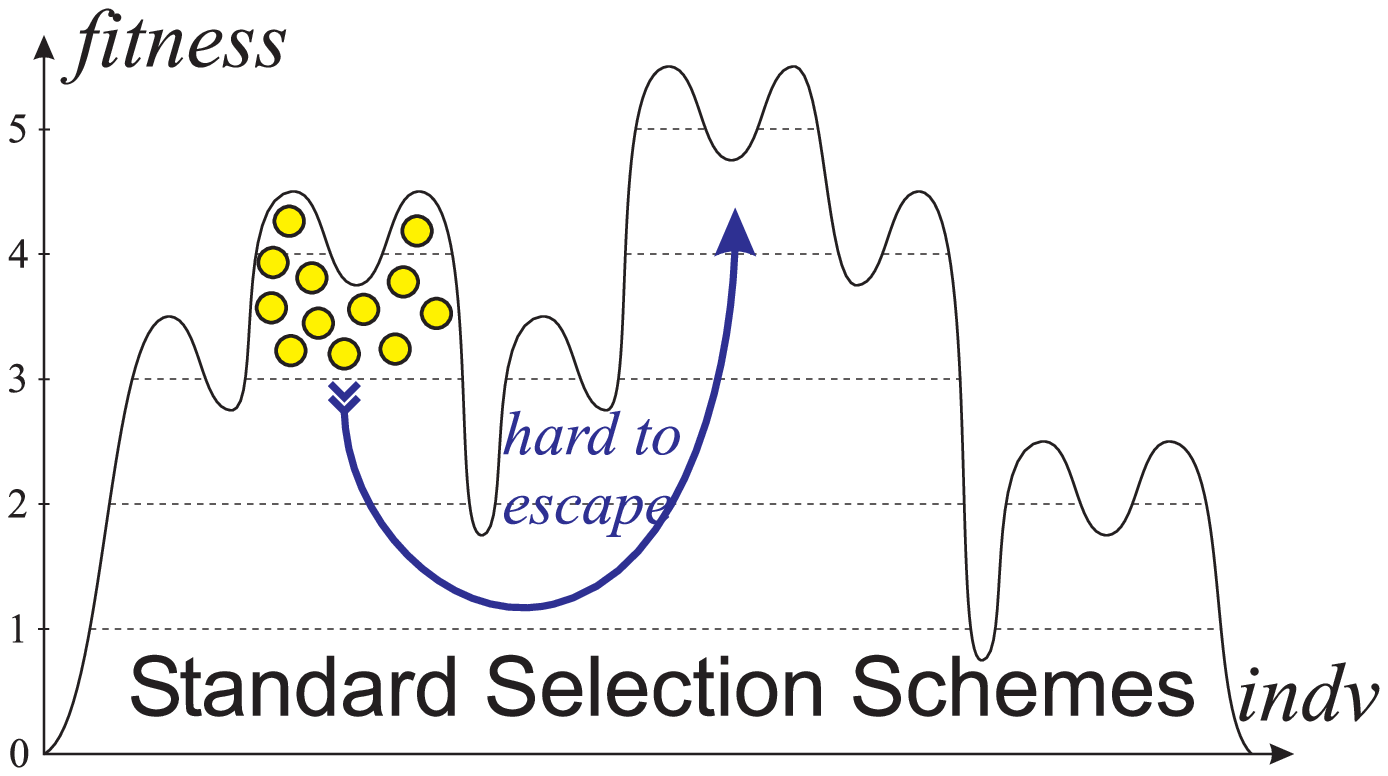}
\caption{\label{figevolve}Evolution of the population under
FUSS versus standard selection schemes (STD): STD may get stuck
in a local optimum if all unfit individuals were eliminated too
quickly. In FUSS, all fitness levels remain occupied with ``free''
drift within and in-between fitness levels, from which new mutants
are steadily created, occasionally leading to further
evolution in a more promising direction.}
\end{figure*}

\emph{Section \ref{secSim}} briefly explains how the relative fitness
of individuals can be used to define a simple metric that provides us
with some indication of the similarity of individuals.
In \emph{Section \ref{secFuss}} we discuss the definition of the
fitness uniform selection scheme for both discrete and continuous
valued fitness functions.  We also outline some of its key theoretical
properties and contrast these with standard selection schemes.
\emph{Section \ref{secJfuss}} details our experimental setup.
\emph{Section \ref{secEx}} examines the performance of FUSS and
tournament selection on an artificially constructed deceptive
optimization problem.  We compare our results to the behavior
predicted in \cite{Hutter:01fuss}.
The performance of FUSS and tournament selection is then examined on a
set of randomly generated integer valued functions in \emph{Section
\ref{secRandFunc}}.
In \emph{Section \ref{secTSP}} we detail the performance of FUSS and
tournament selection on both artificial and real traveling salesman
problems.
In \emph{Section \ref{secSetCover}} we examine the set covering
problem, an NP hard optimization problem which has many real world
applications.
For our final test in \emph{Section \ref{secSAT}} we compare FUSS and
tournament selection on random maximum CNF3 SAT problems and graph
coloring problems which have been expressed in the CNF form.  These
are also NP hard optimization problems.
\emph{Section \ref{secConc}} contains a brief summary of our results and
possible avenues for future research.

%%%%%%%%%%%%%%%%%%%%%%%%%%%%%%%%%%%%%%%%%%%%%%%%%%%%%%%%%%%%%%%
\section{Using Fitness to Measure Similarity}\label{secSim}
%%%%%%%%%%%%%%%%%%%%%%%%%%%%%%%%%%%%%%%%%%%%%%%%%%%%%%%%%%%%%%%

There are many ways to measure the similarity of individuals in a
population.  If the individuals are binary coded one might use the
Hamming distance as a similarity relation. This distance is consistent
with a mutation operator which flips a few bits. It produces
Hamming-similar individuals, but recombination (like crossover) can
produce very dissimilar individuals w.r.t.\ this measure. In any case,
genotypic similarity relations, like the Hamming distance, depend on
the representation of the individuals as binary strings. Individuals
with very dissimilar genomes might actually be functionally
(phenotypically) very similar. For instance, when most bits are unused
(like introns in GP), they can be randomly disturbed without affecting
the property of the individual. For specific problems at hand, it
might be possible to find suitable representation-independent
functional similarity relations. On the other hand, in genetic
programming, for instance, it is in general undecidable, whether two
individuals are functionally similar.

FUSS takes a different approach. The distance between two
individuals $i$ and $j$ with fitness $f(i)$ and $f(j)$ is defined
as
$$
  d(i,j) \;:=\; |f(i)-f(j)|.
$$
The distance is based solely on the fitness function, which is
provided as part of the problem specification.
It is independent of the coding/representation and other problem
details, and of the optimization algorithm (e.g.\ the genetic mutation
and recombination operators), and can trivially be computed from
the fitness values.
If we make the natural assumption that functionally similar
individuals have similar fitness, they are also similar w.r.t.\ the
distance $d$. On the other hand, individuals with very different
coding, and even functionally dissimilar individuals may be
$d$-similar, but we will see that this does not matter. For instance,
individuals from different local optima of equal height are
$d$-similar.

Armed with this simple measure of similarity between individuals we
can now define a selection scheme that aims to preserve diversity in
the population.

%%%%%%%%%%%%%%%%%%%%%%%%%%%%%%%%%%%%%%%%%%%%%%%%%%%%%%%%%%%%%%%
\section{Fitness Uniform Selection Strategy (FUSS)}\label{secFuss}
%%%%%%%%%%%%%%%%%%%%%%%%%%%%%%%%%%%%%%%%%%%%%%%%%%%%%%%%%%%%%%%

The idea behind FUSS is that we should focus the selection pressure
towards fitness levels which have relatively few individuals rather
than on the highest fitness levels.  In this way fitness levels which
are difficult to reach are thoroughly explored and on no fitness level
does the population size decrease towards extinction (see Figure
\ref{figevolve}).  Thus FUSS preserves genetic diversity more actively
than the standard selection schemes which tend to drive the
populations on lower fitness levels to zero.  Moreover, parts of the
fitness space which are interesting, in the sense that they are
difficult to reach, are focused on, rather than easy to reach areas
which are already well represented in the population.  This approach
might seem counter intuitive as we are not even attempting to increase
the average fitness of the population!  The point is that for
optimization problems we are usually only interested in finding a
single individual with the highest possible fitness --- having low
average fitness is not in itself a problem.

\begin{figure}
\centerline{\includegraphics[width=0.9\columnwidth]{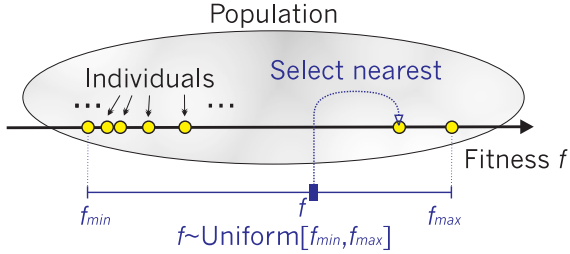}}
\caption{\label{figfuss}If the lowest/highest fitness values
in the current population $P$ are $f_{min/max}$, FUSS selects a
fitness value $f$ uniformly in the interval $[f_{min},f_{max}]$,
then, the individual $i\in P$ with fitness nearest to $f$ is
selected and a copy is added to $P$, possibly after mutation and
recombination.}
\end{figure}

For general real-valued fitness functions FUSS is defined as follows:
A uniform random number is chosen in the interval $[f_{min},f_{max}]$,
where $f_{max}$ and $f_{min}$ are the maximum and minimum fitness
values in the current population. Then the individual with fitness
nearest to this number is chosen (see Figure \ref{figfuss}). If this
is ambiguous one of the nearest individuals is chosen at random.  In
the case of integer valued fitness functions this is equivalent to
selecting a fitness level at random from the set $\{f_{min}, f_{min}
+1, \ldots, f_{max} \}$ and then randomly selecting an individual
within that fitness level if the level is occupied.  If the level is
empty, higher and lower fitness levels are progressively searched
until a non empty level is found at which time a random individual is
selected.

\begin{figure}
\centerline{\includegraphics[width=0.9\columnwidth]{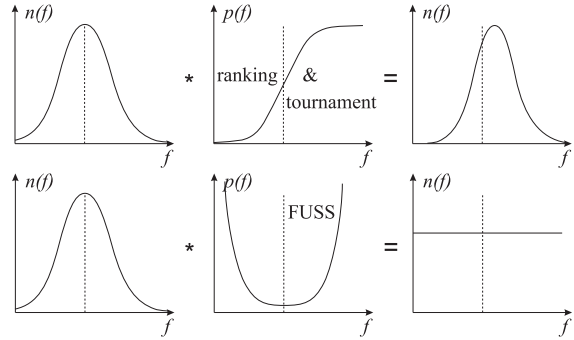}}
\caption{\label{figsel}Effects of ranking=tournament and fitness
uniform (FUSS) selection on the fitness distribution in a generation
based EA.  The left/right diagrams depict fitness distributions
before/after applying the selection schemes depicted in the middle
diagrams.}
\end{figure}

While the probability of selecting each fitness level is equal, the
probability of then selecting a given individual within a fitness level
depends on the population of that level.  For example, if an
individual belongs to a fitness level with 50 members its selection
probability is twice as high as an individual that belongs to a
fitness level with 100 members.  It is easy to see that under a
selection scheme based on the FUSS approach, the proportion of
individuals at each fitness level tends towards the fraction
$\frac{1}{|F|}$, where $|F|$ is the number of fitness levels as
depicted in Figure \ref{figsel}. See \cite{Hutter:01fuss} for a more
detailed description.

While this preserves a greater degree of population diversity than the
standard selection schemes, it comes at the cost of a potential loss
of performance due to the large number of selections from low fitness
levels.  Thus the currently highest parts of the fitness space are now
searched more slowly than under a standard selection scheme.  In the
worst case FUSS will slow the performance of the system down by a
factor of ${|F|}$.  However for significantly deceptive problems the
loss of performance due to becoming stuck in a local optimum for a
long period of time is a much more significant cost than a potential
factor of $|F|$.  It is for these problems that the author of
\cite{Hutter:01fuss} expects the strengths of FUSS to become evident.

At first glance it might appear that there is little pressure on
selecting highly fit individuals under FUSS.  Usually this is not the
case as the most fit individuals in a population are typically quite
rare.  If these individuals start to make up a significant proportion
of the total population this indicates that this part of the space has
been significantly searched and thus is more likely to be an
evolutionary dead end.  In this case FUSS will, by its nature,
automatically move selection pressure away from these highly fit
individuals and focus its search energy on lower fitness levels that
have fewer individuals. In this way the selection intensity varies
dynamically with the evolution of the population.  Clearly this is
preferable to the situation where we must manually fix the selection
pressure for a particular optimization problem in order to prevent the
system from becoming stuck in local optima.

%%%%%%%%%%%%%%%%%%%%%%%%%%%%%%%%%%%%%%%%%%%%%%%%%%%%%%%%%%%%%%%
\section{GA Test System}\label{secJfuss}
%%%%%%%%%%%%%%%%%%%%%%%%%%%%%%%%%%%%%%%%%%%%%%%%%%%%%%%%%%%%%%%

We have implemented a GA test system in Java on a PC running Linux.
The selection schemes include FUSS and the standard tournament
selection scheme.  With tournament selection we randomly pick a group
of individuals and then select the fittest individual from this group.
The size of the group is called the \emph{tournament size} and it is
clear that the larger this group is the more likely we are to select a
highly fit individual from the population.  A tournament size of 2 is
commonly used in practice as this often provides sufficient selection
intensity on the most fit individuals.  In our tests we have used
tournament sizes of 2, 5 and 15 which we will refer to as TOUR2, TOUR5
and TOUR15 respectively.  This should provide some insight into how
different levels of selection intensity affect performance in
different problems.

We have chosen to compare FUSS with tournament selection as this
scheme is simple to understand and implement and is also one of the
most widely used.  Also we consider it to be roughly representative of
other standard selection schemes which favor the fitter individuals in
the population; indeed in the case of tournament size 2 it can be show
that tournament selection is equivalent to the linear ranking
selection scheme \cite[Sec.2.2.4]{Hutter:92cfs}. At some point in the
future we may implement other standard selection schemes to broaden
our comparison, however we expect the performance of these schemes to
be at best comparable to tournament selection when used with a
correctly tuned selection intensity.

The GA model we have chosen is the so called ``steady state'' model as
opposed to the more usual ``generational'' model.  In a generational
GA at each generation we select an entirely new population based on
the old population.  The old population is then simply discarded.
Under the steady state model that we use, individuals are only
selected one at a time: We select an individual, then with a certain
probability we select another and cross the two to produce a new
individual, and then with another probability we mutate the result.
We will refer to the probability of crossing as the \emph{crossover
probability} and the probability of mutating following a cross as the
\emph{mutate probability}.  In the case where no crossover took place
the individual is always mutated to insure that we are not simply
adding a clone of an existing individual into the population.  Finally
an individual must be deleted in order to keep the population size
constant.  How this is done is important as it can bias the population
in a way that is similar to the selection scheme.  We have chosen to
simply delete a random individual from the population which is a
common neutral strategy used in steady state GAs.

The number of generations in a generational GA is roughly equivalent
to the number of iterations in a steady state GA divided by the
population size.  We have used this approximation here when reporting
the number of generations on graphs etc.  Unfortunately the
theoretical understanding of the relationship between the two types of
GA is quite poor.  It has been shown that under the assumption of no
crossover the effective selection intensity using tournament selection
with size 2 is approximately twice as strong under a steady state GA
as it is with a generational GA \cite{Rogers:99}.  As far as we are
aware a similar comparison for systems with crossover has not been
performed, though we would not expect the results to be significantly
different.  While steady state GAs have certain advantages, the fact
that generational GAs are more common means that we may in the future
test FUSS under this model also.

The important free parameters to set for each test are the population
size, and the crossover and mutation probabilities mentioned above.
Our default is to have both the crossover and mutation probabilities
set to 0.5.  For each problem we conducted some preliminary
experiments to establish reasonable settings for these variables.
Often the effect of these variables on performance was not
particularly strong, though it was always worth checking to be sure.
More importantly, the relative performance of the selection schemes
remained quite stable.  For population sizes less than 500 performance
tended to degrade for difficult problems where the potential solution
space was large.  To avoid this our experiments have been performed
with populations of 1,000 individuals or more.  For each test the
parameters were the same for each selection scheme --- indeed the only
difference was which subroutine in the code was used to select
individuals.  This ensures that our comparison was fair.

In order to generate reliable statistics we ran each test multiple
times; typically 20 or 30 times.  From these runs we then calculated
the average performance for each selection scheme.  We also computed
the sample standard deviation and from this the standard error in our
estimate of the mean.  This value was then used to generate the 95\%
confidence intervals which appear on the graphs.

%%%%%%%%%%%%%%%%%%%%%%%%%%%%%%%%%%%%%%%%%%%%%%%%%%%%%%%%%%%%%%%
\section{A Deceptive 2D Problem}\label{secEx}
%%%%%%%%%%%%%%%%%%%%%%%%%%%%%%%%%%%%%%%%%%%%%%%%%%%%%%%%%%%%%%%

The first problem we examine is the simple but highly deceptive 2D
problem for which the performance of FUSS was theoretically
analyzed in \cite{Hutter:01fuss}.  The setup of the test is quite
simple.  The space of individuals is the unit square
$[0,1]\times[0,1]$.  On this space narrow regions $I_1
:= [a,a+\delta] \times [0,1]$ and $I_2 := [0,1] \times
[b,b+\delta]$ for some $a,b,\delta \in [0,1]$ are defined.  Typically
$\delta$ is chosen so that it is much smaller than 1 and thus $I_1$
and $I_2$ do not occupy much of the domain space.  The fitness
function is defined to be,
$$
  f(x,y) = \left\{
  \begin{array}{l}
    1 \quad\mbox{if}\quad (x,y)\in I_1\backslash I_2, \\
    2 \quad\mbox{if}\quad (x,y)\in I_2\backslash I_1, \\
    3 \quad\mbox{if}\quad (x,y)\not\in I_1\cup I_2, \\
    4 \quad\mbox{if}\quad (x,y)\in I_1\cap I_2. \\
  \end{array}\right.
\parbox{2.5cm}
{\hfill \unitlength=0.6mm
%\linethickness{0.4pt}
\begin{picture}(45,45)
\scriptsize
\put(5,5){\vector(0,1){40}}
\put(5,5){\vector(1,0){40}}
\put(20,5){\line(0,1){35}}
\put(25,5){\line(0,1){35}}
\put(40,5){\line(0,1){35}}
\put(5,15){\line(1,0){35}}
\put(5,20){\line(1,0){35}}
\put(5,40){\line(1,0){35}}
\put(22.5,17.5){\makebox(0,0)[cc]{4}}
\put(12.5,30){\makebox(0,0)[cc]{3}}
\put(22.5,30){\makebox(0,0)[cc]{1}}
\put(32.5,30){\makebox(0,0)[cc]{3}}
\put(32.5,10){\makebox(0,0)[cc]{3}}
\put(12.5,10){\makebox(0,0)[cc]{3}}
\put(12.5,17.5){\makebox(0,0)[cc]{2}}
\put(32.5,17.5){\makebox(0,0)[cc]{2}}
\put(22.5,10){\makebox(0,0)[cc]{1}}
\put(44,2.5){\makebox(0,0)[cc]{$x$}}
\put(22.5,2.5){\makebox(0,0)[cc]{$\delta$}}
\put(20,3.5){\makebox(0,0)[cc]{$a$}}
\put(2.5,17.5){\makebox(0,0)[cc]{$\delta$}}
\put(4,14.5){\makebox(0,0)[cc]{$b$}}
\put(40,3){\makebox(0,0)[cc]{1}}
\put(3.5,40){\makebox(0,0)[cc]{1}}
\put(2.5,44){\makebox(0,0)[cc]{$y$}}
\put(22.5,42.5){\makebox(0,0)[cc]{$f(x,y)$}}
\end{picture}
}
$$
The example has sort of an XOR structure, which is
hard for many optimizers.

For this problem we set up the mutation operator to randomly set
either the $x$ or $y$ position of an individual and the crossover to
take the $x$ position from one individual and the $y$ position from
another to produce an offspring.

Under these operators this is a very deceptive and difficult
optimization problem.  The size of the domain for which the function
is maximized is just $\delta^2$ which is very small for small values
of $\delta$.  Moreover the local maxima at fitness level 3 covers most
of the space and the only way to reach the global maximum is by
leaving this local maxima and exploring the space of individuals with
lower fitness value of 1 or 2.  For such a problem FUSS should in
theory perform much better than either random search or more standard
selection schemes.

\begin{figure}
\centerline{\includegraphics[width=\columnwidth]{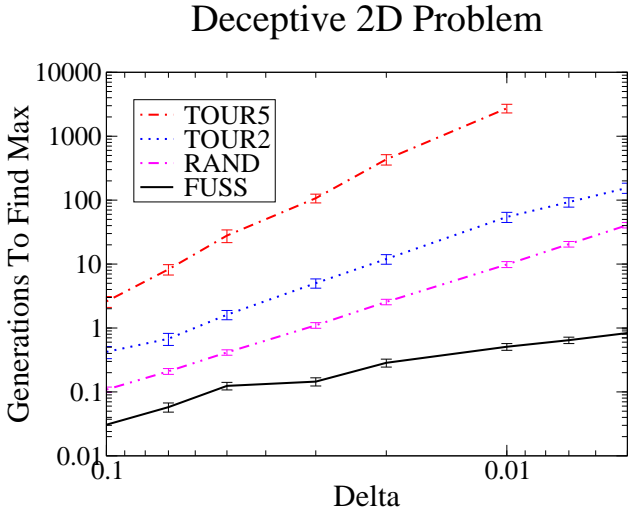}}
\caption{\label{SimpleProb}As predicted FUSS scales significantly
better than both tournament selection and random search for this
problem.  Also, increasing selection pressure in tournament selection
(TOUR2 vs. TOUR5) degraded performance.}
\end{figure}

For this test we set the maximum population size to 10,000 and ran
each scheme for each delta value 20 times.  With a steady state GA it
is usual to start with a full population of random individuals.
However for this particular problem we reduced the initial population
size down to just 10 in order to avoid the effect of doing a large
random search when we created the initial population and thereby
distorting the scaling.  Usually this might create difficulties due to
the poor genetic diversity in the initial population.  However due to
the fact that any individual can mutate to any other in just two steps
this is not a problem in this situation.  Initial tests indicated that
reducing the crossover probability from 0.5 to 0.25 improved the
performance slightly and so we have used this setting.  For comparison
random search (RAND) was also implemented.  The results of these tests
appear in Figure~\ref{SimpleProb}.

As expected higher selection pressure on the most fit individuals is
clearly a disadvantage for this problem.  With low selection pressure
(TOUR2) tournament selection performs slightly worse than random
search while with medium selection pressure (TOUR5) performance was in
the order of 20 times slower than random search.  With high selection
pressure (TOUR15) the test became infeasible to compute.  Our results
confirm the theoretical scaling factors of $1\over\delta^2$ for RAND
and TOUR2, and $1\over\delta$ for FUSS, as predicted in
\cite{Hutter:01fuss}.

%%%%%%%%%%%%%%%%%%%%%%%%%%%%%%%%%%%%%%%%%%%%%%%%%%%%%%%%%%%%%%%
\section{Random Functions}\label{secRandFunc}
%%%%%%%%%%%%%%%%%%%%%%%%%%%%%%%%%%%%%%%%%%%%%%%%%%%%%%%%%%%%%%%

\begin{figure*}
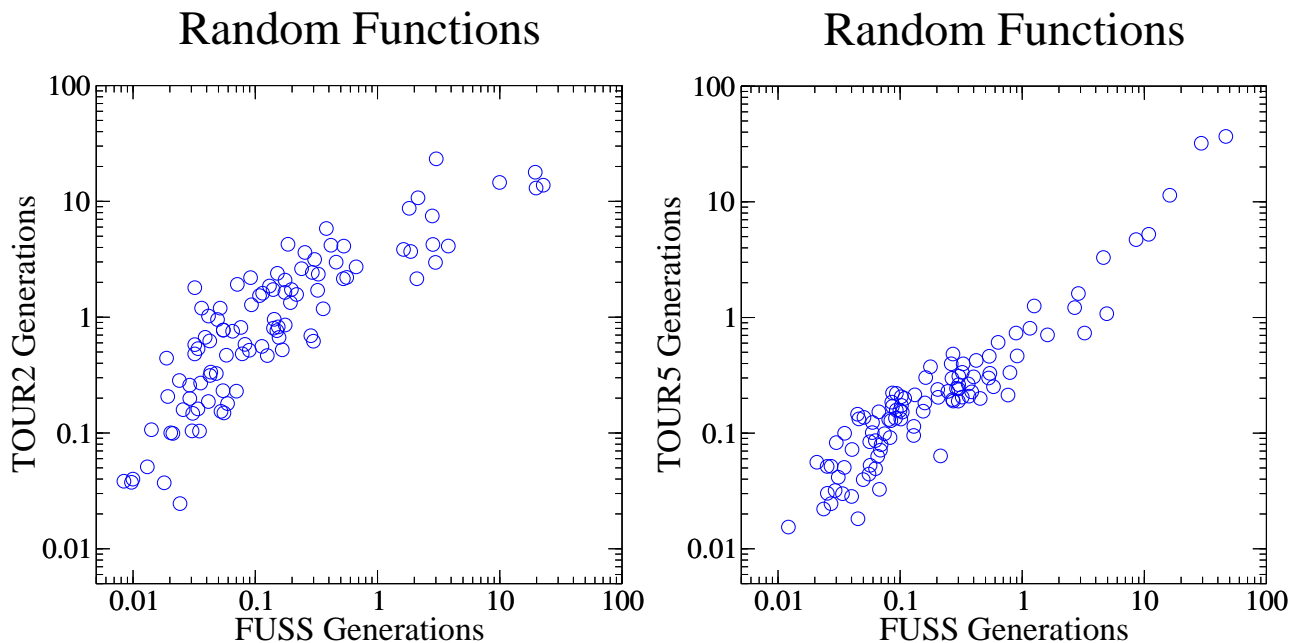

\includegraphics[width=0.5\textwidth]{RandomFunct05S2.eps}
\includegraphics[width=0.5\textwidth]{RandomFunct09S5.eps}
\caption{\label{RandFunct1}\label{RandFunct2} FUSS finds the
global maximum typically ten times faster than tournament
selection with low selection pressure (left diagram). With
increased selection pressure, tournament reaches, but does not
pass the performance of FUSS (right diagram).}
\end{figure*}

In order to gain a better understanding of how FUSS performs relative
to tournament selection in more general problem settings we tested the
selection schemes on a set of randomly generated functions.  In this
case the domain of each function was the 4 dimensional hyper cube
$[0,1]^4$.  To create each random function we randomly generated 16
cuboids of 4 dimensions inside the domain space.  The function value
of a point inside the domain space was then taken to be the number of
random cuboids that contained the point.  Thus, depending on where the
random cuboids where, the range of the function could be anything from
$\{0,1\}$ to $\{0,1,\ldots,16\}$.  This process of building up
functions using cuboids allowed the functions to be quite complex and
multi modal while still keeping some rough continuity.  In order to
make the optimization problem a little more tractable we limited the
width of the cuboids in any one dimension to be in the range
$[0.2,1]$.  While this limited the minimum size of each random cuboid,
two or more cuboids could still form arbitrarily small intersections
and thus the domain region in which a function achieves its maximal
value could still be extremely small.

In the first test we generated 100 random functions and precomputed
each function's global maximum value by using the cuboid position
information used to construct the function.  For each function we then
ran both FUSS and TOUR2 10 times and computed for each the average
number of generations needed to find the global maximum.  This
produced 100 data points corresponding to the 100 random functions.

We first tested TOUR2 as we expected the problem to be relatively
deceptive and thus higher selection pressure would be a disadvantage.
The population size for these tests was set at 10,000.  The results
are plotted in Figure~\ref{RandFunct1}.

We see that FUSS typically manages to find the global maximum 2 to 20
times faster than TOUR2, with 10 being about average.  We then
compared FUSS with TOUR5.  The results of this test are plotted in
Figure~\ref{RandFunct2}.

Interestingly the performance of tournament selection improved to the
extent that it was then roughly equivalent to FUSS.  Typically a
tournament size of 2 is sufficient selection pressure for most
problems.  This performance improvement due to increased selection
pressure indicates that tournament selection wasn't becoming
significantly stuck in local optima and thus these random function
problems where not as deceptive as we had anticipated.  We increased
selection pressure further by testing TOUR15, but no further
performance gains were to be had.

While the strength of FUSS is in dealing with very difficult and
deceptive optimization problems, this result demonstrates that even
for problems where greater selection pressure is an advantage the
performance of FUSS can remain competitive.  FUSS also had the
advantage that no parameter tuning was required in order to achieve
optimum performance for this problem.

%%%%%%%%%%%%%%%%%%%%%%%%%%%%%%%%%%%%%%%%%%%%%%%%%%%%%%%%%%%%%%%
\section{Traveling Salesman Problem}\label{secTSP}
%%%%%%%%%%%%%%%%%%%%%%%%%%%%%%%%%%%%%%%%%%%%%%%%%%%%%%%%%%%%%%%

\begin{figure*}
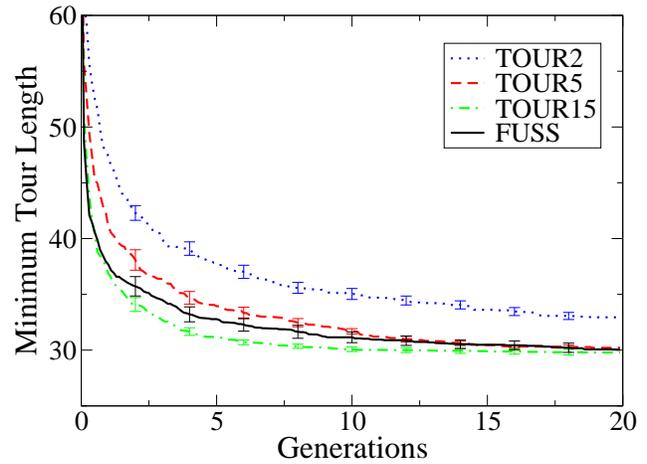

\includegraphics[width=0.5\textwidth]{DTSP.eps}
\includegraphics[width=0.5\textwidth]{TSPsahara.eps}
\caption{\label{DTSP}\label{Sahara} FUSS performs well in both
artificial and real TSP problems.  For the random TSP problem FUSS
converges much faster than TOUR2 but also manages to avoid becoming
stuck in a local optimum like TOUR15.  In the real TSP problem FUSS
again has close to the optimal selection intensity.}
\end{figure*}

To find the shortest Hamiltonian cycle (path) in a graph of $N$
vertices (cities) connected by edges of certain lengths is a difficult
optimization problem. In the following we present preliminary results
of a simple evolutionary TSP optimizer with standard selection (here
tournament selection) and with FUSS.  There are highly specialized
(evolutionary) algorithms finding paths less than one percent longer
than the optimal path for up to $10^7$ cities
\cite{Lin:73,Martin:96,Johnson:97,Applegate:00}.  Whether FUSS could
further improve these algorithms will be studied elsewhere. Here, we
are just interested in the performance of FUSS compared to tournament
selection on a difficult optimization problem that has real world
applications.

The mutation and crossover operators we used were quite simple.
Mutation was done by simply switching the position of two of the
cities in the solution.  For crossover we used the common partial
mapped crossover technique \cite{Goldberg:85}.

The first test was carried out on a set of TSP problems with random
distance matrices.  There were 50 TSP problems in total each with 20
cities.  The distance between any two cities was chosen uniformly from
the interval $[0,1]$.  This is a particularly deceptive form of the
TSP problem as the usual triangle inequality relation does not hold.
For example, the distance between cities $A$ and $B$ might be $0.1$,
between cities $B$ and $C$ $0.2$, and yet the distance between $A$ and
$C$ might be $0.8$.  The problem still has some structure though as
efficient partial solutions tend to be useful building blocks for
efficient complete tours.  For this test we used a population size of
5,000 and the default mutation and crossover rates of 0.5.  The
results appear in Figure~\ref{DTSP}.

We see here that the selection intensity with TOUR2 is too low for the
system to converge in a reasonable number of generations.  On the
other hand the selection intensity under TOUR15 is too high and causes
the system to become stuck in a local optimum.  TOUR5 has about the
correct selection intensity for this problem.  FUSS outperforms both
TOUR2 and TOUR15 and is very close to TOUR5 at the end of the run.

We also tested the system on a number of real TSP problems based on
the location of real cities from various countries around the world
\cite{Applegate:03}.  For these tests the population size was set at
5,000.  Based on experimentation we increased the crossover
probability to 1.0 and the probability of mutation was reduced to 0.2
for better performance.  The results were averaged over a total of 5
runs.  The results for the ``Sahara'' dataset are shown in
Figure~\ref{Sahara}.

Here we see that a higher level of selection intensity is appropriate.
FUSS again performs significantly better than TOUR2 and also
a little better than TOUR5.  At the end of the run FUSS has converged
to the same level as both TOUR5 and TOUR15 which is again a positive
result for FUSS.

We tested the system on a number of other datasets under various other
parameter settings for population size, rate of mutation and crossover
etc. and obtained similar results.  Nevertheless a fuller analysis
comparing other possible mutation and crossover operations and
parameters settings will need to be done before more substantive
conclusions are possible.

%%%%%%%%%%%%%%%%%%%%%%%%%%%%%%%%%%%%%%%%%%%%%%%%%%%%%%%%%%%%%%%
\section{Set Covering Problem}\label{secSetCover}
%%%%%%%%%%%%%%%%%%%%%%%%%%%%%%%%%%%%%%%%%%%%%%%%%%%%%%%%%%%%%%%

The set covering problem (SCP) is a reasonably well known NP-complete
optimization problem with many real world applications.  Let $M \in
\{0,1\}^{m \times n}$ be a binary valued matrix and let $c_j > 0$ for
$j \in \{1, \ldots n \}$ be the cost of column $j$.  The goal is to
find a subset of the columns such that the cost is minimized.  Define
$x_j = 1$ if column $j$ is in our solution and 0 otherwise.  We can
then express the cost of this solution as $\sum_{j=1}^n c_j x_j$
subject to the condition that $\sum_{j=1}^n m_{ij} x_j \geq 1$ for $i
\in \{1, \ldots m\}$.

Our system of representation, mutation operators and crossover follow
that used by Beasley \cite{Beasley:96}. We compared the performance of
FUSS with tournament selection on a number of standard test problems
\cite{Beasley:03}. For these tests we set the population size to
5,000, crossover probability to 1.0, the mutation probability to 0.5 and
averaged the performance of the systems over 30 runs on each problem.

\begin{figure*}
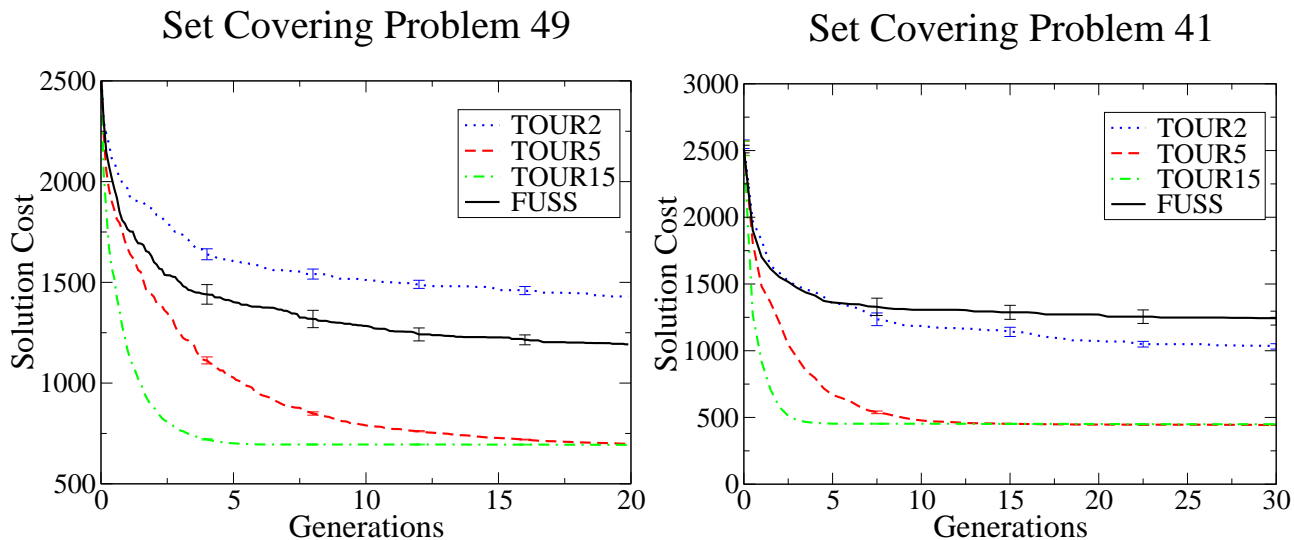

\includegraphics[width=0.5\textwidth]{scp49.eps}
\includegraphics[width=0.5\textwidth]{scp41.eps}
\caption{\label{scp49}\label{scp41} In SCP49 FUSS converges more
quickly than TOUR2 but is still too slow.  SCP41 is an easy problem as
TOUR15 find the optimum very quickly however for some reason FUSS
becomes stuck far from the optimum.}
\end{figure*}

The results in Figure~\ref{scp49} were based on the ``SCP49'' and
``SCP41'' datasets.  Here the performance of FUSS is less impressive.
For SCP49 FUSS performs better than TOUR2 however the rate of
convergence is still too low.  SCP41 is an easy problem with TOUR15
converging in just 4 generations.  Nevertheless FUSS is converging
very slowly, if at all.  It is interesting that FUSS performs poorly
on this relatively easy problem when its performance was strong on
more difficult problems such as random TSP and the deceptive 2D
problem presented earlier.  We will look more closely into the reasons
for this in the next section.

%%%%%%%%%%%%%%%%%%%%%%%%%%%%%%%%%%%%%%%%%%%%%%%%%%%%%%%%%%%%%%%
\section{Maximum CNF3 SAT}\label{secSAT}
%%%%%%%%%%%%%%%%%%%%%%%%%%%%%%%%%%%%%%%%%%%%%%%%%%%%%%%%%%%%%%%

Maximum CNF3 SAT is a well known NP hard optimization problem
\cite{Crescenzi:04} that has been extensively studied.  A three
literal conjunctive normal form (CNF) logical equation is a boolean
equation that consists of a conjunction of clauses where each clause
contains a disjunction of three literals.  So for example, $(a \lor b
\lor \lnot c) \land ( a \lor \lnot e \lor f)$ is a CNF3 expression.
The goal in the maximum CNF3 SAT problem is to find an instantiation
of the variables such that the maximum number of clauses evaluate to
true.  Thus for the above equation if $a = F$, $b = T$, $c = T$, $e =
T$, and $f = F$ then just one clause evaluates to true and thus this
instantiation gets a score of one.  Achieving significant results in
this area would be difficult and this is not our aim; we are simply
using this problem as a test to compare FUSS and tournament
selection.

Our test problems have been taken from the SATLIB collection of SAT
benchmark tests \cite{Hoos:00}.  The first test was performed on 30
instances of randomly generated CNF3 forumlae with 150 variables and
645 clauses which are all known to be satisfiable.  The second test
was performed on 30 instances of ``flat'' 3 colorable graph coloring
problems with 50 vertices and 115 edges which have been expressed in
CNF form.  The graph coloring problems have a slightly different
structure as the clauses contain either 2 or 3 literals.

\begin{figure*}
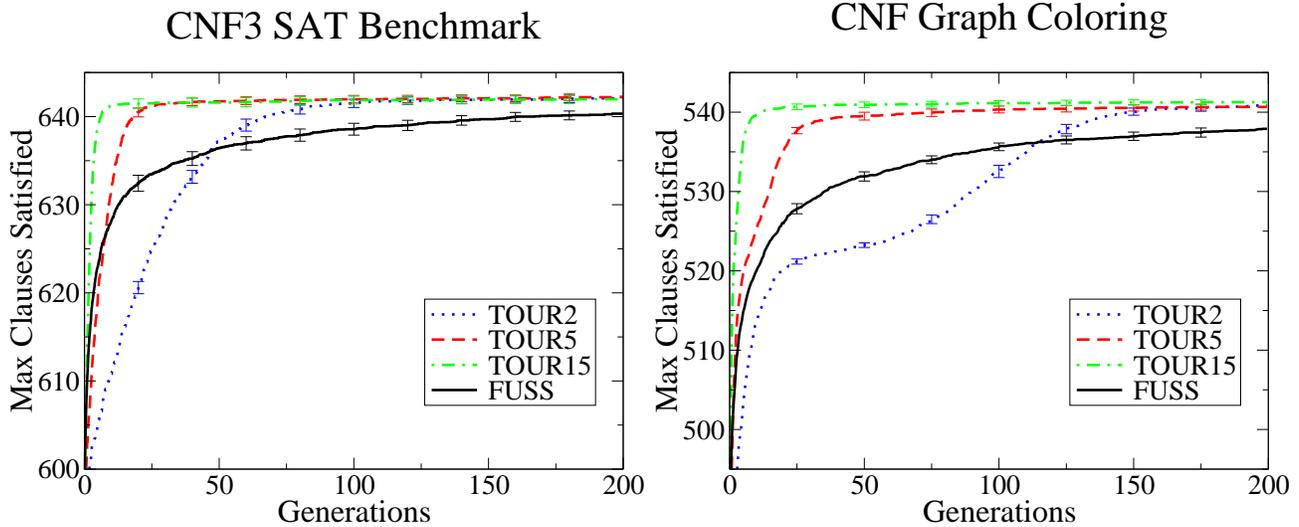

\includegraphics[width=0.5\textwidth]{CNF-150-SAT-subset.eps}
\includegraphics[width=0.5\textwidth]{CNF-flat40-subset.eps}
\caption{\label{CNF3SAT}In the CNF3 SAT benchmark with 150 variables and
645 clauses the performance of FUSS was slightly below the performance
of tournament selection.  The graph coloring problem shows a similar
result.}
\end{figure*}

Our mutation operator simply flips one boolean variable and the
crossover operator forms a new individual by randomly selecting for
each variable which parent's state to take.  The population size was
set to 10,000 and the crossover and mutation probabilities were left
at the default setting of 0.5.  The test was run 30 times for each
selection method.  The results for both tests appear in
Figure~\ref{CNF3SAT}.

In both tests we see that the maximum fitness under FUSS initially
climbs very rapidly.  Closer inspection showed that it climbs even
more rapidly than TOUR15 for the first 0.5 generations.  This
indicates that FUSS has an extremely high selection intensity to start
with, much higher than even TOUR15.  After this period FUSS starts to
slow down.  It appears to become either stuck in a local optimum or
the selection intensity falls dramatically, either way, it is then
easily passed by the tournament selection schemes.  We also tested the
system with controlled backbone CNF problems from the same set of
benchmark tests and obtained similar results.

We can explain this behavior by considering a simple example.
Consider a situation where there is a large number of individuals in a
small band of fitness levels, say 10,000 with fitness values ranging
from 50 to 70.  Add to this population one individual with a fitness
value of 73.  Thus the total fitness range is now 24.  Whenever FUSS
picks a random point from 72 to 73 inclusive this single individual
with maximal fitness will be selected.  That is, the probability that
the single fittest individual will be selected is 2/24 = 0.083.  Now
compare this to TOUR15, a selection scheme with high selection
intensity.  Under TOUR15 the probability that the fittest individual
is selected is the same as the probability that it is picked for the
sample of 15 elements used for the tournament, that is, 15/10000 =
0.0015.  Thus we can see that in this simple example the probability of
selecting the fittest individual under FUSS is over 50 times higher
than what it is under TOUR15.  This effectively gives FUSS an
extremely high selection intensity and would likely result in a very
rapidly rising maximal fitness value.  If a mutant derived from our
highly fit individual had a fitness value higher than 73 then the
situation would become much more extreme causing the system to rapidly
explore this evolutionary path and fill the higher fitness levels with
many highly related individuals in the process.

Once a high level of fitness is reached and further progress becomes
difficult the distribution of individuals across the fitness range
balances out.  When this happens the selection probability for
individuals at the highest fitness levels converges towards
$\frac{1}{|P|}$ where $|P|$ is the size of the population.  Thus the
selection intensity becomes very low, much lower than under TOUR15.
This explains why FUSS becomes stuck after its initial rapid rise in
maximal fitness.

Further experiments have been carried out to test whether these
difficulties are responsible for the performance problems we have
seen.  While FUSS is suited for problems where it is difficult to
directly measure and thus control diversity, in the CNF problems we
are able measure diversity quite easily by computing hamming distance.
Doing so reveals that the diversity in the total population remains
very high under FUSS over the evolution of the system, much higher
than under the tournament selection schemes.  This is what we would
expect to see given that FUSS maintains a broad set of both fit and
unfit individuals in the population.  However if we look at the
genetic diversity in the top 10\% of the population we see that
diversity under FUSS falls very rapidly and is generally significantly
worse than under the tournament selection scheme.  Thus while we have
succeeded in preserving diversity in the population as a whole, among
the fittest individuals in the population diversity is actually rather
poor.  This is consistent with the scenario described above where FUSS
tends to over exploit a very small number of fit individuals in the
population.

%%%%%%%%%%%%%%%%%%%%%%%%%%%%%%%%%%%%%%%%%%%%%%%%%%%%%%%%%%%%%%%
\section{Conclusions \& Future Research Directions}\label{secConc}
%%%%%%%%%%%%%%%%%%%%%%%%%%%%%%%%%%%%%%%%%%%%%%%%%%%%%%%%%%%%%%%

Theoretical analysis suggests that FUSS should be able to outperform
standard selection schemes in some situations, in particular on highly
deceptive optimization problems.  Our results for a deceptive 2D
optimization problem and for TSP problems confirm this.  However we
have also observed cases where FUSS has performance difficulties.
Further analysis indicates that this is due to the greedy nature of
FUSS selection in the early stages of the system's evolution.  While
total genetic diversity was very strong, diversity among the most fit
individuals was poor due to the nature of our selection scheme.  This
suggests that while fitness can be used to control diversity, our
current method of doing so is inadequate.  We are currently
investigating alternates to FUSS which achieve diversity across
fitness levels while not exploiting small groups of fit individuals
too heavily in the process.  Our results so far have been encouraging
with diversity being strong both in the population as a whole and
among fit individuals.

\end{document}